\documentclass{article}

\usepackage{microtype}
\usepackage{graphicx}
\usepackage{subfigure}
\usepackage{booktabs} 

\usepackage{hyperref}

\usepackage[accepted]{icml2025}

\usepackage{amsmath}
\usepackage{amssymb}
\usepackage{mathtools}
\usepackage{amsthm}

\usepackage[capitalize,noabbrev]{cleveref}

\theoremstyle{plain}

\theoremstyle{definition}

\theoremstyle{remark}

\usepackage[textsize=tiny]{todonotes}

\usepackage{makecell}
\usepackage{array}
\usepackage{enumitem}
\usepackage[normalem]{ulem}
\usepackage[most]{tcolorbox}
\usepackage{verbatim}
\usepackage{listings}

\icmltitlerunning{Submission and Formatting Instructions for ICML 2025 Workshop on Assessing World Models}

\begin{document}

\twocolumn[
\icmltitle{FRED: Financial Retrieval-Enhanced Detection and Editing of Hallucinations in Language Models}



\icmlsetsymbol{equal}{*}

\begin{icmlauthorlist}
\icmlauthor{Likun Tan}{equal,comp}
\icmlauthor{Kuan-Wei Huang}{equal,comp}
\icmlauthor{Kevin Wu}{comp}
\end{icmlauthorlist}

\icmlaffiliation{comp}{Pegasi AI, New York, USA}

\icmlcorrespondingauthor{Kevin Wu}{kevin@usepegasi.com}

\icmlkeywords{Machine Learning, ICML}

\vskip 0.3in
]



\printAffiliationsAndNotice{\icmlEqualContribution} 

\begin{abstract}
Hallucinations in large language models pose a critical challenge for applications requiring factual reliability, particularly in high-stakes domains such as finance. This work presents an effective approach for detecting and editing factually incorrect content in model-generated responses based on the provided context. Given a user-defined domain-specific error taxonomy, we construct a synthetic dataset by inserting tagged errors into financial question-answering corpora and then fine-tune four language models, Phi-4, Phi-4-mini, Qwen3-4B, and Qwen3-14B, to detect and edit these factual inaccuracies. Our best-performing model, fine-tuned Phi-4, achieves an 8\% improvement in binary F1 score and a 30\% gain in overall detection performance compared to OpenAI-o3. Notably, our fine-tuned Phi-4-mini model, despite having only 4 billion parameters, maintains competitive performance with just a 2\% drop in binary detection and a 0.1\% decline in overall detection compared to OpenAI-o3. Our work provides a practical solution for detecting and editing factual inconsistencies in financial text generation while introducing a generalizable framework that can enhance the trustworthiness and alignment of large language models across diverse applications beyond finance. Our code and data are available at \url{https://github.com/pegasi-ai/shield}.
\end{abstract}

\section{Introduction}

Large Language Models (LLMs) such as GPT-4 and PaLM have demonstrated impressive capabilities across a wide range of natural language processing (NLP) tasks. However, a persistent challenge remains: these models often generate hallucinations---plausible-sounding but factually incorrect content that is not grounded in the provided input or external knowledge sources~\citep{ji2023survey, mishra2024fine}. To address hallucination in language models, Retrieval-Augmented Generation (RAG)~\citep{lewis2020retrieval} has emerged as a widely adopted framework. By grounding responses in relevant evidence, RAG reduces—but does not eliminate—the occurrence of factual errors. Consequently, it becomes essential to develop methods that can assess the factual consistency of generated text with respect to the retrieved context (i.e., hallucination detection) and propose appropriate corrections when inconsistencies are found (i.e., hallucination editing).

Hallucination detection has been explored through a range of techniques. One line of work focuses on fine-tuning encoder-based models, as demonstrated in LUNA~\citep{belyi2024luna} and LettuceDetect~\citep{kovacs2025lettucedetect}. Another approach leverages LLM-as-a-judge strategies, where LLMs are used to directly assess factual consistency, as in DeepEval~\citep{deepeval2025} and Glider~\citep{deshpande2024glider}. More recent methods aim to quantify the model’s reliance on retrieved evidence, such as REFIND~\citep{lee2025refind} and ReDeEP~\citep{sun2024redeep}. Meanwhile, hallucination editing has recently gained attention as a complementary task to hallucination detection. GENAUDIT~\citep{krishna2024genaudit} identifies unsupported spans in LLM outputs and suggests evidence-grounded edits, while FAVA~\citep{min2023factscore} and PFME~\citep{deng2024pfme} perform fine-grained correction by leveraging span-level annotations to classify and revise erroneous content. 

In this work, we introduce a novel framework for detecting and editing factual inaccuracies in model-generated answers with respect to a retrieved context in financial RAG systems. To this end, we make the following contributions:
\begin{itemize}[left=0pt]
\item{\bf Domain-specific Error Taxonomy}: We define a taxonomy of factual error types in the financial domain, informed by linguistic patterns and reasoning errors observed in financial question answering tasks. This taxonomy includes categories such as numerical miscalculations, incorrect entity references, temporal inconsistencies, and relation mismatches.
\item{\bf Synthetic Dataset Construction}: Leveraging this taxonomy, we introduce tagged hallucinations into answers from FinQA~\citep{chen2021finqa} and TAT-QA~\citep{zhu2021tat} - two benchmark datasets requiring multi-step numerical reasoning over financial tables and text. Errors are inserted via controlled perturbations, enabling supervised training of hallucination-aware models.
\item{\bf End-to-End Hallucination Editing}: Our fine-tuned small language models (SLMs) perform span-level hallucination detection and correction, identifying incorrect phrases and replacing them with contextually grounded edits. This setup enables interpretable and traceable corrections suitable for finance-specific applications.
\end{itemize}

\section{Data Curation}
\label{sec:data}

We utilize two publicly available datasets, FinQA and TAT-QA, both sourced from RagBench~\citep{friel2024ragbench} to train and evaluate our model. These datasets comprise financial question-answering tasks that require multi-step numerical reasoning over textual contexts. Each example includes a \textit{document} consisting of retrieved evidence, a \textit{question} related to the document, and a corresponding \textit{response}. We retained only the examples in which the \textit{response} contains information that is grounded in the provided \textit{document}. To evaluate the model's editing capability, we construct training data comprising a passage with factual errors (obtained from the original \textit{response}) and its corrected version as the target output. We define six types of factual errors - \texttt{Temporal}, \texttt{Numerical}, \texttt{Entity}, \texttt{Relation}, \texttt{Contradictory} and \texttt{Unverifiable} - specifically tailored to the kinds of mistakes that may arise when answering questions based on earnings calls. Detailed definitions of these error types are provided in Appendix~\ref{app:tags}.
The data curation process comprises multiple steps to generate the final target output. Before detailing the procedure, we introduce the input, intermediate representations, and final outputs by defining key terminologies and providing illustrative examples.

{\bf Original Passage}: The original \textit{response} from datasets. \\
Example: \texttt{The annual interest expense for \\entergy louisiana incurred from the \\series first mortgage bonds due September 2018 is \$19.5 million.} 

{\bf Tagged Passage}: An annotated version of the passage with structured tags indicating factual errors. Editable errors (e.g., \texttt{Temporal}, \texttt{Numerical}, \texttt{Entity}, \texttt{Relation}) are annotated using \texttt{<mark>} and \texttt{<delete>} to highlight the inserted error and the original span. \\
Example: \texttt{The annual interest expense for \\entergy louisiana incurred from the \\series first mortgage bonds due \\<temporal><delete>September 2018\\</delete><mark>August 2008</mark>\\</temporal> is \$19.5 million. <unverifiable>The bond proceeds were \\primarily used to fund confidential\\ environmental initiatives.</unverifiable>}

{\bf Erroneous Passage}: The corrupted version of the passage, obtained by removing all tags and retaining the inserted errors. \\ 
Example: \texttt{The annual interest expense for \\entergy louisiana incurred from the \\series first mortgage bonds due August 2008 is \$19.5 million. The bond proceeds were primarily used to fund confidential environmental initiatives.}

{\bf Target Output}: The corrected version of the erroneous passage with structured tags reintroduced to indicate errors and provide corrections. \\
Example: \texttt{The annual interest expense for \\entergy louisiana incurred from the\\ series first mortgage bonds due <temporal><mark>September 2018</mark>\\<delete>August 2008</delete></temporal> is \$19.5 million. <unverifiable>\\The bond proceeds were primarily used to fund confidential environmental initiatives.</unverifiable>}

With the terminologies defined above, we follow a three-stage pipeline to curate training data: 
\begin{enumerate}[left=0pt]
\item Error Insertion: We adopt the systematic error insertion strategies outlined in Appendix~\ref{app:error_insertion} to generate \textbf{tagged passage}, ensuring controlled and diverse error generation across different linguistic and factual dimensions.

\item Filtering and correction: Given that model-generated outputs are not guaranteed to be fully consistent or valid, we perform a comprehensive quality check based on four criteria introduced in Appendix~\ref{app:preliminary}: \textbf{Incorrect Type}, \textbf{Identical Text}, \textbf{Invalid Format}, and \textbf{Inconsistent Content}. Instances with \textit{unfixable} issues (e.g., \textbf{Invalid Format} or \textbf{Inconsistent Content}) are discarded, while \textit{fixable} issues (e.g., \textbf{Incorrect Type} or \textbf{Identical Text}) are programmatically corrected to ensure high-quality training data.

\item Training Data Preparation:  To construct training instances, we need \textbf{erroneous passage} that is within the structured prompts alongside reference context, and the corresponding \textbf{target output} consisting of edited versions of these passages, reflecting corrections to the hallucinated content. Both the \textbf{erroneous passage} and \textbf{target output} are derived via post-processing of the \textbf{tagged passsage}, enabling the creation of aligned input-output pairs for subsequent fine-tuning.
\end{enumerate}

A diagram representation of the procedures is given in Figure~\ref{fig:data_curation}. We constructed two synthetic training datasets for our experiments. The first dataset consists of ~8K examples, with ~3K samples derived from FinQA and ~5K from TATQA. Additionally, we prepared a larger dataset comprising ~36K training examples, including ~11K from FinQA and ~25K from TATQA. The distribution of error types in this 36K dataset is presented in Table~\ref{tab:error_type_distribution}. We split each dataset into training and validation sets using a 95:5 ratio. For evaluation, we use a separate test set derived from FAVA and FinQA+TATQA.

\begin{table}[htbp]
\centering
\caption{Distribution of Error Types across FinQA and TATQA}
\begin{tabular}{lccc}
\toprule
 & \multicolumn{3}{c}{\textbf{Percentage (\%)}} \\
\cmidrule(lr){2-4}
\textbf{Type} & \textbf{FinQA} & \textbf{TATQA} & \textbf{Total} \\
\midrule
Hallucinated     & 69.8\% & 66.5\% & 67.5\% \\
Non-hallucinated & 30.2\% & 33.5\% & 32.5\% \\
Numerical Errors    & 23.2\% & 17.9\% & 20.0\% \\
Temporal Errors     & 31.9\% & 30.0\% & 30.8\% \\
Entity Errors       & 13.1\% & 13.9\% & 13.6\% \\
Relation Errors     & 7.4\% & 7.9\% & 7.7\% \\
Contradictory Statements & 16.4\% & 20.1\% & 18.6\% \\
Unverifiable Statements   & 8.0\% & 10.1\% & 9.2\% \\
\bottomrule
\end{tabular}
\label{tab:error_type_distribution}
\end{table}

\section{Experiments}\label{sec:experiment}

Following~\citet{mishra2024fine}, we evaluate two key aspects of factual verification: hallucination detection and hallucination editing. Specifically, we investigate the effectiveness of fine-tuned SLMs and benchmark their performance against LLMs. Towards this, we fine-tuned four SLMs-—\texttt{Phi-4}, \texttt{Phi-4 Mini}, \texttt{Qwen3-4B}, and \texttt{Qwen3-14B}. The training details can be seen in Appendix~\ref{app:model}.

\subsection{Experiments for Hallucination Detection} 

\paragraph{Comparison with baseline models on FAVA.} We begin our experiments using the FAVA dataset. This dataset comprises approximately 35,000 synthetic instances, each consisting of a pair of hallucinated inputs and their corresponding corrected outputs. The raw data is drawn from Wikipedia-based open-domain content, spanning topics such as science, history, politics, and technology. Designed for factuality verification, the dataset is well-suited for evaluating and training models on tasks in general field.

For our baseline models, we selected \texttt{GPT-4.1-mini} and \texttt{o3}. \texttt{GPT-4.1-mini} is a lightweight variant of OpenAI’s GPT-4.1 model, offering strong language understanding with lower computational overhead—making it ideal for resource-constrained settings. In contrast, \texttt{o3} (also known as GPT-4-o or Omni) is OpenAI’s flagship multimodal model, featuring enhanced reasoning, factual accuracy, and performance across modalities. It underpins the latest versions of ChatGPT and API services.

Table~\ref{tab:fava_f1} presents the F1-scores of various editors across fine-grained hallucination categories in the factuality detection task. Phi4\textsuperscript{*} achieves the highest overall performance, significantly surpassing other models across all error types. Notably, it demonstrates exceptional performance in detecting \texttt{Invented} (93.6) and \texttt{Unverifiable} (90.3) hallucinations, yielding the highest overall F1-score (79.8) and binary detection score (92.1). \texttt{o3} ranks second, exhibiting strong performance on \texttt{Entity} hallucinations (67.2) and binary detection (89.7). These results highlight the efficacy of small, fine-tuned models such as Phi4\textsuperscript{*} in identifying hallucinated content across diverse factual error categories, even outperforming advanced proprietary models like \texttt{o3}. More results on precision and recall metrics can be found in Table~\ref{tab:fava_precision} and Table~\ref{tab:fava_recall} in Appendix~\ref{app:detection}

\begin{table}[htbp]
\centering
\caption{Detection Performance (F1-score) on FAVA}
\begin{tabular}{lcccccccc}
\toprule
\textbf{Editor} & \textbf{Ent.} & \textbf{Rel.} & \textbf{Con.} & \textbf{Inv.} & \textbf{Sub.} & \textbf{Unv.} & \textbf{Ov.} & \textbf{Bi.} \\
\midrule
{\footnotesize GPT-4.1 mini} &  34.3&  39.7&  53.1&  50.0&  \underline{71.7}&  37.1&  54.6& 82.3\\
o3     &  \textbf{67.2} &  \underline{40.0}&  \underline{65.5}&  \underline{72.5}&  59.6&  \underline{69.4} &  \underline{69.8} & \underline{89.7}\\
Phi4-mini$^*$ & 42.4  & 37.8 & 52.5 & 52.2 & 61.8 & 44.8 & 57.8 & 88.6 \\
Phi4$^*$ & \underline{54.9} & \textbf{62.9} & \textbf{81.8} & \textbf{93.6} & \textbf{85.1} & \textbf{90.3} & \textbf{79.8} & \textbf{92.1} \\
\bottomrule
\end{tabular}
\label{tab:fava_f1}
\\\footnotesize * are fine-tuned models.
\\\footnotesize Ent. = Entity, Rel. = Relation, Con. = Contradictory, Inv. = Invented, Sub. = Subjective, Unv. = Unverifiable, Ov. = Overall, Bi. = Binary.
\end{table}

\paragraph{Comparison with baseline models on FinQA+TATQA.} Building on our initial analysis using the FAVA dataset, we shift our focus to evaluating the factual editing capabilities of SLMs in the financial domain. For this purpose, we use FinQA and TATQA as our primary training and evaluation datasets. We conducted experiments using two training datasets. The first consists of 8k examples, comprising 3k samples from FinQA and 5k from TATQA. The second is a larger dataset with 36k examples, including 11k from FinQA and 25k from TATQA. Additionally, we introduced two new models—\texttt{QWEN3-4b} and \texttt{QWEN3-14b}—open-weight dense models from the latest Qwen family. These models feature enhanced reasoning capabilities, performing detailed, step-by-step reasoning to tackle complex tasks.

For baseline models, we selected \texttt{GPT-4.1-mini} and \texttt{o3}, consistent with our earlier experiments. To align with our predefined error types, we modified the prompt accordingly to guide the generation process. The prompt used for the baseline models is given in Appendix~\ref{app:prompt}.

We report the F1 scores in Table~\ref{tab:finqa_f1}. Results of precision and recall are provided in Table~\ref{tab:finqa_precision} and~\ref{tab:finqa_recall}, respectively, in the Appendix~\ref{app:detection}. Phi4-36k\textsuperscript{*} achieves the highest performance across most categories, with top F1 scores in \texttt{Numerical} (86.0), \texttt{Temporal} (93.3), and \texttt{Entity}(92.1), as well as the best Overall score (93.8) and binary detection score (97.5). Notably, in the binary detection task, all the 14b models (including Phi4-36k\textsuperscript{*}, Phi4-8k\textsuperscript{*} and QWEN3-14b-36k\textsuperscript{*}) outperform \texttt{o3}, while smaller models like Phi4-mini-36k\textsuperscript{*} and QWEN3-4b-36k\textsuperscript{*} deliver comparable performance. In addition, results from Phi4-36k\textsuperscript{*} and Phi4-mini-36k\textsuperscript{*} are generally better than their counterparts with 8k training datasets, indicating that increasing the amount of training data consistently improves model performance, suggesting further gains are achievable with additional data.

\begin{table}[htbp]
\centering
\small
\caption{Detection Performance (F1-score) on FinQA+TATQA}
\begin{tabular}{lcccccccc}
\toprule
\textbf{Editor} & \textbf{Num.} & \textbf{Tem.} & \textbf{Ent.} & \textbf{Rel.} & \textbf{Con.} & \textbf{Unv.} & \textbf{Ov.} & \textbf{Bi.} \\
\midrule
GPT-4.1 mini & 23.5 & 24.6 & 19.8 & 55.2 & 21.5 & 42.1 & 46.0 & 77.8 \\
o3     &  52.2&  83.8&  63.8&  46.6&  29.4&  78.9&  71.9& 90.3\\
Phi4-mini-8k$^*$      & 29.1 & 51.9 & 42.0 & 38.6 & 70.4 & 76.1 & 63.9 & 83.8 \\
Phi4-mini-36k$^*$  & 51.1 & 66.2 & 40.0 & 40.6 & 84.1 & 76.2 & 72.0 & 88.3 \\
Phi4-8k$^*$       & \underline{71.4} & \underline{85.7} & \underline{79.5} & \underline{80.8} & \textbf{95.0} & \textbf{96.6} & \underline{89.9} & \underline{96.7} \\
Phi4-36k$^*$     & \textbf{86.0} & \textbf{93.3} & \textbf{92.1} & \textbf{88.1} & \underline{94.3} & \underline{94.5} & \textbf{93.8} & \textbf{97.5} \\
QWEN3-4b-36k$^*$   & 48.2 & 58.8 & 54.3 & 50.7 & 82.6 & 76.2 & 72.4 & 89.7 \\
QWEN3-14b-36k$^*$ & 57.7 & 72.0 & 59.7 & 45.5 & 86.5 & 83.7 & 77.6 & 91.1 \\
\bottomrule
\end{tabular}
\label{tab:finqa_f1}
\\\footnotesize * are fine-tuned models.
\\\footnotesize Num. = Numerical, Tem. = Temporal, Ent. = Entity, Rel. = Relation, Con. = Contradictory, Unv. = Unverifiable, Ov. = Overall, Bi. = Binary.
\end{table}






\subsection{Experiments for Hallucination Editing}

Hallucination editing enables a more fine-grained evaluation of the factual accuracy of generated text. In this study, we utilize \texttt{FActScoreLite}~\citep{FactScoreLite}, a streamlined and modular reimplementation of the \texttt{FactScore} metric, specifically designed for detailed factuality assessment in natural language generation. The framework consists of two primary components: (i) the \texttt{AtomicFactGenerator}, which decomposes long-form model outputs into discrete atomic factual statements, and (ii) the \texttt{FactScorer}, which evaluates the factual correctness of each atomic statement against a reference document. For our purposes, we employ only the \texttt{FactScorer} module to assess the factual alignment of model-generated text with respect to the source content. While the original implementation of \texttt{FActScoreLite} supports only \texttt{gpt-4-turbo-preview} as the evaluation backend, we extend the framework to incorporate additional model options, including \texttt{gpt-4-turbo}, \texttt{o3-mini}, and \texttt{LLaMA-Scout} for a more comprehensive evaluation.

Tables~\ref{tab:editing_fava} and~\ref{tab:editing_finqa} report editing results on the FAVA and FinQA+TATQA datasets, respectively. ``No Edit'' denotes the \textbf{erroneous passage} prior to any correction. On the FAVA dataset, \texttt{o3} achieves the highest performance, indicating strong capabilities in hallucination correction, followed closely by Phi4\textsuperscript{*}. Interestingly, although Phi4\textsuperscript{*} outperforms \texttt{o3} in detection tasks, \texttt{o3} shows superior performance in editing, likely due to better fact-checking abilities despite less consistent formatting compliance. On FinQA+TATQA, Phi4-36k\textsuperscript{*} slightly outperforms \texttt{o3} when scored with \texttt{gpt-4-turbo}, while \texttt{o3} maintains a consistent lead across the remaining evaluation settings. Another note is that although \texttt{GPT-4.1 mini} demonstrates significant improvement on the FAVA editing task, it shows minimal improvement on FinQA+TATQA, indicating its relatively limited ability for numerical reasoning compared to the other models. In contrast, both fine-tuned \texttt{Phi4-mini} and \texttt{Phi4} exhibit consistent performance improvements across both FAVA and FinQA+TATQA, showcasing their strong proficiency in numerical reasoning tasks.

\begin{table}[htbp]
\centering
\caption{Editing Results on FAVA}
\label{tab:editing_fava}
\begin{tabular}{lccc}
\toprule
 & \multicolumn{3}{c}{\textbf{Model}} \\
\cmidrule(lr){2-4}
\textbf{Editor} & \textbf{gpt-4-turbo} & \textbf{o3-mini} & \textbf{Llama-scout} \\
\midrule
No Edit   & 34.4 & 34.5 & 41.2 \\
GPT-4.1 mini & 66.9 & 64.2 & 69.6 \\
o3    & \textbf{92.6} & \textbf{95.3} & \textbf{89.9} \\
Phi4-mini$^*$  & 68.2 & 68.2 & 73.7 \\
Phi4$^*$       & \underline{85.1} & \underline{81.1} & \underline{82.4} \\
\bottomrule
\end{tabular}
\\\footnotesize * are fine-tuned models.
\end{table}

\begin{table}[htbp]
\centering
\caption{Editing Results on FinQA+TATQA}
\label{tab:editing_finqa}
\begin{tabular}{lccc}
\toprule
 & \multicolumn{3}{c}{\textbf{Model}} \\
\cmidrule(lr){2-4}
\textbf{Editor} & \textbf{gpt-4-turbo} & \textbf{o3-mini} & \textbf{Llama-scout} \\
\midrule
No Edit   & 40.1  & 40.4 & 43.8 \\
GPT-4.1 mini & 40.1 & 41.1 & 41.8 \\
o3    & \underline{91.0} & \textbf{94.5} & \textbf{90.4}\\
Phi4-mini-8k$^*$     & 73.6 & 71.2 & 69.9 \\
Phi4-mini-36k$^*$    & 78.4 & 69.2 & 73.6  \\
Phi4-8k$^*$       & 87.3 & 82.9 & 83.2 \\
Phi4-36k$^*$     & \textbf{91.4}  & \underline{84.6} & \underline{86.0} \\
QWEN3-4b-36k$^*$     & 74.0  & 78.1  & 74.3 \\
QWEN3-14b-36k$^*$    & 72.9  & 82.5  & 76.7 \\
\bottomrule
\end{tabular}
\\\footnotesize * are fine-tuned models.
\end{table}

\section{Conclusion}\label{sec:conclude}

We present a suite of state-of-the-art SLMs fine-tuned to detect and correct fine-grained hallucinations based on grounded context in the financial domain. Experimental results demonstrate that our models outperform existing approaches in both detection and editing tasks, particularly in terms of overall accuracy and binary detection accuracy, while operating under limited computational resources. 

\section*{Acknowledgements}

We thank Neo for supporting this research through their startup accelerator program. Their contribution played a crucial role in enabling the development and evaluation of our models.

\nocite{langley00}

\bibliography{main}
\bibliographystyle{icml2025}

\newpage
\appendix
\onecolumn

\section{Related Work}
\label{app:related}

\paragraph{Detection and Editing in Context-Grounded LMs.}  
With the growing adoption of Retrieval-Augmented Generation (RAG) systems, evaluating contextual hallucinations has become increasingly important. Recent advancements in this area include post-hoc detection methods~\citep{manakul2023selfcheckgpt}, contrastive verification techniques~\citep{min2023factscore}, fine-grained hallucination detection frameworks~\citep{mishra2024fine, deng2024pfme}, and evidence-grounded editing approaches~\citep{krishna2024genaudit}. Our work builds upon the fine-grained hallucination detection paradigm proposed by~\citep{mishra2024fine}, extending it to the financial domain through the introduction of a domain-specific error taxonomy and an emphasis on practical correction of factual inconsistencies.

\paragraph{Financial Question and Answering.}  
Question answering in the financial domain requires reasoning over both unstructured text and structured tabular data. Several benchmark datasets have been proposed to support numerical reasoning tasks, such as FinQA~\citep{chen2021finqa} and TAT-QA~\citep{zhu2021tat}. State-of-the-art approaches leverage LLMs that are instruction-tuned or domain-adapted to financial corpora, such as BloombergGPT~\citep{wu2023bloomberggpt}, FinMA~\citep{xie2023pixiu} and FinGPT~\citep{wang2023fingpt}. Program generation technique~\citep{chen2021finqa} and graph-structured method ~\citep{barry2025graphrag} have also been employed to improve interpretability and robustness in table-based reasoning. In parallel, multimodal models like Donut~\citep{kim2021donut} and LayoutLMv3~\citep{huang2022layoutlmv3} have been adapted to process complex financial PDFs that blend text, tables, and charts. Our work aims at investigating contextual hallucinations in RAG systems and evaluating the effectiveness of SLMs for fine-grained hallucination annotation with minimal editing.

\section{Definition of Fine-grained Hallucinations}
\label{app:tags}

The design of our error tag taxonomy builds upon the fine-grained hallucination framework introduced by~\citet{mishra2024fine}, which categorizes hallucinations into six distinct types: \texttt{Entity}, \texttt{Relation}, \texttt{Contradictory}, \texttt{Invented}, \texttt{Subjective}, and \texttt{Unverifiable}. This classification provides a structured basis for the detection and correction of factual inconsistencies, contributing to improvements in both model accuracy and response reliability. However, the taxonomy proposed by~\citet{mishra2024fine} exhibits limitations, particularly with regard to the categories of \texttt{Invented}, \texttt{Subjective} and \texttt{Unverifiable}, which are inherently subjective and lack clearly defined linguistic criteria. To address these shortcomings, we consolidate these three categories into a unified \texttt{Unverifiable} class. In addition, we introduce two new categories—\texttt{Temporal} and \texttt{Numerical}—to better accommodate the types of errors frequently observed in financial-domain applications, thus enhancing the domain specificity and practical applicability of the taxonomy. The details of our definition of error tags are shown in Table~\ref{tab:tags}.

\begin{table}[htbp]
 \caption{Tag Definition}
  \centering
  \begin{tabular}{>{\raggedright\arraybackslash}p{3cm}>{\raggedright\arraybackslash}p{10cm}}
    \toprule
    \textbf{Tag Type}     & \textbf{Definition}    \\
    \midrule
    Temporal & \makecell[tl]{\makecell[{{p{10cm}}}]{A temporal error arises when a statement contains incorrect or mismatched time-related information, such as dates, timeframes, fiscal years, or ordering of events. \\ 
    Example: The total amount outstanding in \textcolor{red}{\sout{2010}} \textcolor{green}{2019} is \$ 576.2 million (342.9 + 78.5 + 154.8). }} \\
    Numerical & \makecell[tl]{\makecell[{{p{10cm}}}]{A numerical error involves incorrect or imprecise quantities, percentages, ratios, or other numerical values present in financial data or reasoning. \\ 
    Example: The before tax charge related to adopting Fin No. 47 as of December 31, 2005, was \textcolor{red}{\sout{\$25}} \textcolor{green}{\$19} million.}} \\
    Entity & \makecell[tl]{\makecell[{{p{10cm}}}]{An entity error occurs when a company, organization, geographic location, product name, or financial instrument is incorrectly referenced. \\ 
    Example: In 2014, the net change in \textcolor{red}{\sout{revenue recognition}} \textcolor{green}{tax positions} was an increase of \$17,290.}} \\
    Relation& \makecell[tl]{\makecell[{{p{10cm}}}]{A relation error refers to a misrepresentation of the relationship between entities or financial concepts, such as ownership, causality, comparison, or attribution.\\
    Example: The earnings from service operations \textcolor{red}{\sout{decreased}} \textcolor{green}{increased} from \$32.8 million in 2000 to \$35.1 million in 2001.}} \\
    Contradictory& \makecell[tl]{\makecell[{{p{10cm}}}]{A contradictory error exists when a statement conflicts with the information in the given context or with another part of the response.\\
    Example: The company was profitable in Q4 2022. (Context: The company reported a net loss in Q4 2022.) }}\\
    Unverifiable& \makecell[tl]{\makecell[{{p{10cm}}}]{An unverifiable error includes hallucinated or speculative content that cannot be confirmed or grounded in the provided context or authoritative sources. This also includes vague generalizations or invented data. \\
    Example: The company plans to acquire a fintech startup next quarter. }}\\
    \bottomrule
  \end{tabular}
  \label{tab:tags}
\end{table}

\section{Data Curation}\label{app:data}

\subsection{Preliminary Investigation of Models}\label{app:preliminary} 

To evaluate the robustness of different language models in the context of error insertion, we selected a diverse set of candidate models that represent both proprietary and open-source architectures. Specifically, we include the following:
\paragraph{GPT-family models:} \texttt{gpt-3.5-turbo}, \texttt{gpt-4-turbo}, and \texttt{gpt-4.1-mini}, which are proprietary models developed widely by OpenAI, known for their high performance on a range of natural language understanding tasks.
\paragraph{LLaMA-based models:} \texttt{LLaMA3-70B} and \texttt{LLaMA-4-Maverick}, representing both classic and recent advances in open-weight foundation models. These models are optimized for both capability and computational efficiency.
\paragraph{Gemma family:} \texttt{Gemma2-9B-IT}, an instruction-tuned model that balances performance and cost efficiency, featuring a lightweight architecture optimized for instruction following.

These models were evaluated by prompting them to generate passages containing the error types defined in Table~\ref{tab:tags}, following a controlled error-insertion setup. During the analysis, we observed a variety of systematic errors, which we categorized into the following types:
\begin{itemize}[left=0pt]
\item {\bf Incorrect Type:} The semantic label assigned to a marked span does not accurately reflect the nature of the content. For \\instance, in the example: \texttt{Total Net Cost in <entity><delete>2018</delete><mark>2017</mark>\\</entity> = <numerical><delete>\$3,495 million</delete><mark>\$3,395 million\\</mark></numerical>}, the tag \texttt{<entity>} should have been \texttt{<temporal>} to correctly represent the type of modification. This type of mislabeling was observed across multiple models and reflects a general weakness in fine-grained error-type classification.
\item {\bf Identical Text:} In this case, the model incorrectly generates an edit where the marked and deleted spans are identical, suggesting an unnecessary or meaningless modification. For example: \texttt{The net income as at June 30, 2019 is <relation><delete>\$271,885</delete><mark>\$271,885</mark></relation>}. This error, although rare, was observed in a generation from \texttt{gpt-3.5-turbo}, and should be considered critical as it undermines the purpose of controlled error simulation.
\item {\bf Invalid Format:} The structure of the generated tags violates the expected schema, most commonly through illegal nesting of tags. An illustrative example is: \texttt{<contradictory>The depreciation and amortization expense included as a charge to income was the same for the years ended <temporal>\\<delete>December 31, 2019</delete><mark>June 30, 2018</mark></temporal> and 2018.</contradictory>}. Such structural violations were predominantly observed in outputs from LLaMA-based models.
\item {\bf Inconsistent Content:} The edited passage is inconsistent with the original input, such that the original cannot be reconstructed by simply removing the inserted error markers and texts. Consider the following example:
\begin{itemize}
    \item Original passage: \texttt{Therefore, 2018 has a higher total value of property and \\equipment.} 
    \item Tagged passage: \texttt{Therefore, <contradictory>2019</contradictory> <relation>\\<delete>has</delete><mark>does not have</mark></relation> a higher total value of property and equipment.}
\end{itemize} 
Here, the logical relationship between entities is altered beyond the scope of explicit edits, creating an irrecoverable semantic mismatch. 
\end{itemize}

We further categorize the first two error types, \textbf{Incorrect Type} and \textbf{Identical Text}, as \emph{fixable}, as they can potentially be corrected through post-processing or rule-based validation. In contrast, the latter two categories, \textbf{Invalid Format} and \textbf{Inconsistent Content}, are deemed \emph{unfixable}, since the inserted error tags introduce structural or semantic inconsistencies that are difficult to resolve through any automatic post-processing methods. We manually evaluated ten examples and summarized the errors in the candidate models in Table~\ref{tab:error_insertions}. Among all evaluated models, \texttt{GPT-3.5-turbo}, \texttt{GPT-4}, and \texttt{Gemma2-9B-IT} produce the fewest \emph{unfixable} errors, making them suitable candidates for synthetic training data generation. 

\begin{table}[htbp]
\centering
\caption{Investigation of Models on Error Insertion}
\begin{tabular}{lccccc}
\toprule
 & \multicolumn{2}{c}{\textbf{Fixable}} & \multicolumn{2}{c}{\textbf{Unfixable}} & \\
\cmidrule(lr){2-5} 
\textbf{Model} & \textbf{Inc. Typ.} & \textbf{Ide. Tex.} & \textbf{Inv. For.} & \textbf{Inc. Con.} & \textbf{Tot. Unf.} \\
\midrule
gpt-3.5-turbo     & 2 & 1 & 0 & 1 & 1\\
gpt-4-turbo       & 2 & 0 & 0 & 1 & 1 \\
gpt-4.1-mini      & 1 & 0 & 0 & 3 & 3\\
Llama3-70b        & 0 & 0 & 1 & 2 & 3\\
llama-4-maverick  & 1 & 0 & 2 & 5 & 7\\
gemma2-9b-it      & 1 & 0 & 0 & 1 & 1\\
\bottomrule
\end{tabular}
\label{tab:error_insertions}
\\\footnotesize Inc. Typ. = Incorrect Type, Ide. Tex. = Identical Text, Inv. For. = Invalid Format, Inc. Con. = Inconsistent Content, Tot. Unf. = Total Unfixable.
\end{table}

\subsection{Strategies for error insertion}\label{app:error_insertion} 

Based on our preliminary investigation, systematic errors during error insertion are inevitable. To generate high-quality synthetic data for hallucination detection and editing, we adopt the following strategies for controlled error insertion:

\paragraph{Model Selection for Error Injection.} To enrich and diversify error patterns, we employ a range of language models—\texttt{GPT-3.5-turbo}, \texttt{GPT-4}, and \texttt{Gemma2-9B-IT}—which produce the fewest \emph{unfixable} errors. This multi-model approach enables a more robust assessment of error type manifestations, minimizing dependence on model-specific biases.

\paragraph{Controlled and Systematic Variation of Error Types.}  Error injection is guided by a dynamic control mechanism that scales the number of inserted errors in proportion to the token length of each passage, ensuring balanced perturbation across both short and long texts. During generation, error types are stochastically sampled from a predefined taxonomy to avoid over-representation of specific categories. This approach promotes a diverse and representative error distribution, enhancing the effectiveness of downstream training and evaluation.

\paragraph{Prompt Diversification with Few-shot Demonstrations.} To mitigate prompt overfitting and encourage lexical and structural diversity, we present the model with varied few-shot exemplars labeled by error type. These exemplars reflect realistic, domain-specific scenarios (e.g., in finance) and guide the model toward generating plausible yet systematically flawed responses. This strategy improves the generalizability of error patterns while preserving alignment with predefined error taxonomies.

\subsection{Synthetic Data Generation}

\begin{figure}[H] 
\vskip -0.2in
\centering
\includegraphics[width=1.0\textwidth]{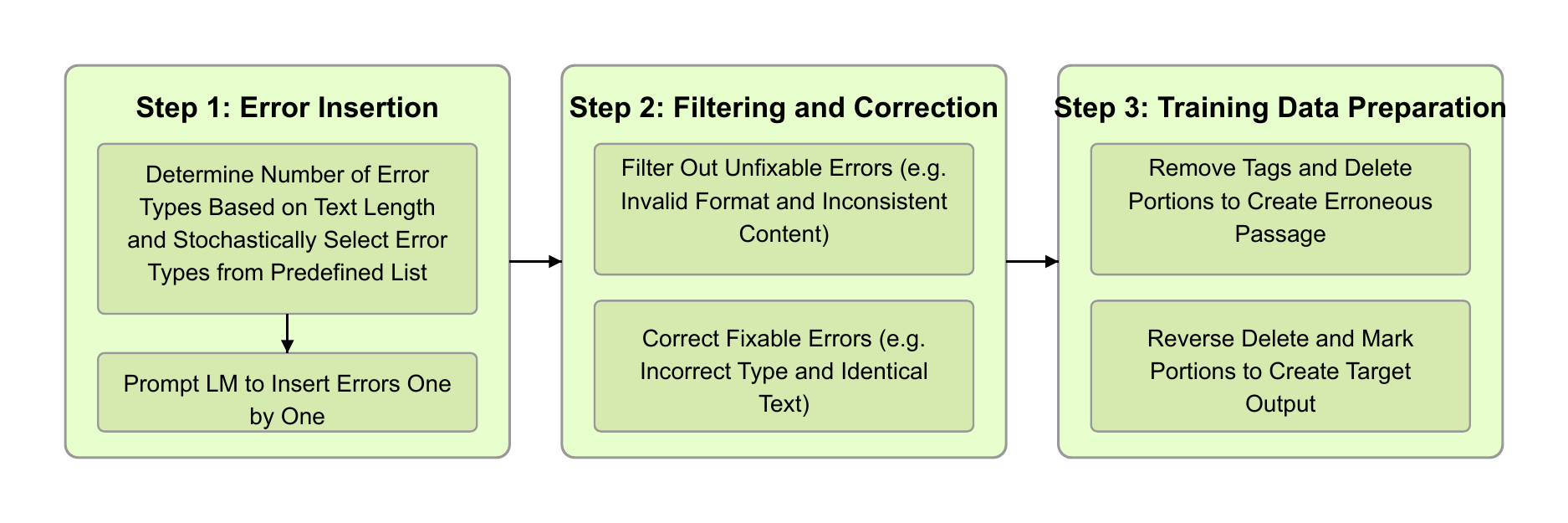}
\vskip -0.2in
\caption{Overview of high-quality synthetic data generation process.}
\label{fig:data_curation}
\end{figure}

Figure~\ref{fig:data_curation} shows the three steps for synthetic data generation.

\section{Model}
\label{app:model}

We fine-tuned four SLMs—\texttt{Phi-4}, \texttt{Phi-4 Mini}, \texttt{Qwen3-4B}, and \texttt{Qwen3-14B}—using LoRA with rank 16, targeting both attention and MLP projection layers (\texttt{q\_proj}, \texttt{k\_proj}, \texttt{v\_proj}, \texttt{o\_proj}, \texttt{gate\_proj}, \texttt{up\_proj}, \texttt{down\_proj}). We set \texttt{lora\_alpha} to 16 and disabled dropout to ensure training stability. For \texttt{Phi-4}, \texttt{Qwen3-4B}, and \texttt{Qwen3-14B}, we employed the Unsloth framework~\citep{unsloth} for efficient 4-bit fine-tuning with a context window of 8192 tokens and enabled gradient checkpointing to reduce memory usage. Since Unsloth does not support \texttt{Phi-4 Mini}, we resort to the Hugging Face ecosystem~\citep{phi4mini2025} with matched hyperparameters. All models were fine-tuned using \texttt{SFTTrainer} with mixed-precision training and linear learning rate scheduling. Training was conducted over two epochs with periodic evaluation. Full training details are summarized in Table~\ref{tab:training}.

We trained the models on two datasets: (1) the publicly available FAVA dataset, which contains ~30k annotated examples of erroneous passages with corresponding tagged corrections, and (2) our own composite dataset derived from FinQA and TATQA, reformulated for an error-tagging task to improve factual consistency and editing capabilities. Each sample included supporting references (e.g., table entries and/or factual statements) and an \textbf{erroneous passage} that may contain factual inaccuracies. The \textbf{target output} was a minimally revised version of the passage, with errors explicitly identified and corrected. This setup trains the models to detect factual inconsistencies and perform token-level, context-aware edits grounded in supporting evidence. For inference, we followed the same data curation procedure to generate the \textbf{erroneous passage} and \textbf{target output}. The \textbf{erroneous passage} was integrated with supporting references to construct a structured prompt, while the \textbf{target output} was reserved for evaluation.

All training experiments were conducted using an NVIDIA A100 GPU with 40 GB of memory, accessed through Google Colab. The training hyperparametes are shown in Table~\ref{tab:training}.

\begin{table}[htbp]
\centering
\caption{Training hyperparameters.}
\begin{tabular}{ccccccc}
\toprule
Precision & Max. Seq. Len. & Epochs & Optim. & LR   & Weight Decay & Warmup Ratio\\
\midrule
bf16 & 8192 &  2  & adamw\_8bit & 2e-5   &  0.01 &  0.03 \\
\bottomrule
\end{tabular}
\label{tab:training}
\end{table}

\section{Detection Task}

\subsection{More Results of Hallucination Detection}\label{app:detection}

\paragraph{Overall Precision and Recall on FAVA.} Table~\ref{tab:fava_precision} and Table~\ref{tab:fava_recall} present the precision and recall metrics on FAVA datasets across the models. Consistent with F1-score results, Phi4\textsuperscript{*} achieves the highest scores on both metrics, while o3 ranks second overall. For binary detection, recall is consistently higher than precision across all models, suggesting a tendency to over-edit—even when the input is factually correct. Additionally, Phi4\textsuperscript{*} exhibits an interesting trade-off: it has lower precision than recall for Entity errors but higher precision than recall for Relation errors. This discrepancy may stem from the ambiguous boundary between entity and relation hallucination categories.

\begin{table}[htbp]
\centering
\caption{Detection Performance (Precision) on FAVA}
\begin{tabular}{lcccccccc}
\toprule
\textbf{Editor} & \textbf{Ent.} & \textbf{Rel.} & \textbf{Con.} & \textbf{Inv.} & \textbf{Sub.} & \textbf{Unv.} & \textbf{Ov.} & \textbf{Bi.} \\
\midrule
GPT-4.1 mini &  22.9 &  34.8 &  48.6 & 67.9 & \underline{82.5} & \textbf{92.9} & 48.7 & 69.9 \\
o3     &  \textbf{64.6} &  \underline{53.3} & \underline{69.2} & \underline{76.7} &  47.2&  63.2&  66.4& 81.3\\
Phi4-mini-8k$^*$ & 38.9 & 31.2 & 51.7 & 54.5 & 58.6 & 43.3 & 53.9 &  \underline{81.5} \\
Phi4-8k$^*$   & \underline{41.7} & \textbf{71.8} & \textbf{86.5} & \textbf{95.7} & \textbf{87.8} & \underline{89.5} & \textbf{75.7} & \textbf{86.1} \\
\bottomrule
\end{tabular}
\label{tab:fava_precision}
\\\footnotesize * are fine-tuned models.
\\\footnotesize Ent. = Entity, Rel. = Relation, Con. = Contradictory, Inv. = Invented, Sub. = Subjective, Unv. = Unverifiable, Ov. = Overall, Bi. = Binary.
\end{table}

\begin{table}[htbp]
\centering
\caption{Detection Performance (Recall) on FAVA}
\begin{tabular}{lcccccccc}
\toprule
\textbf{Editor} & \textbf{Ent.} & \textbf{Rel.} & \textbf{Con.} & \textbf{Inv.} & \textbf{Sub.} & \textbf{Unv.} & \textbf{Ov.} & \textbf{Bi.} \\
\midrule
GPT-4.1 mini & 68.3 & 46.0 & 58.6 & 39.6 & 63.5 & 23.2 & 62.0 & \textbf{1.0} \\
o3     &  \underline{70.0} &  32.0& \underline{62.1} &  \underline{68.8} &  \underline{80.8} &  \underline{76.8} & \underline{73.6} & \textbf{1.0} \\
Phi4-mini-8k$^*$  & 46.7 & \underline{48.0} & 53.4 & 50.0 & 65.4 & 46.4 & 62.3 & 97.0 \\
Phi4-8k$^*$       & \textbf{80.0} & \textbf{56.0} & \textbf{77.6} & \textbf{91.7} & \textbf{82.7} & \textbf{91.1} & \textbf{84.4} & \underline{99.0} \\
\bottomrule
\end{tabular}
\label{tab:fava_recall}
\\\footnotesize * are fine-tuned models.
\\\footnotesize Ent. = Entity, Rel. = Relation, Con. = Contradictory, Inv. = Invented, Sub. = Subjective, Unv. = Unverifiable, Ov. = Overall, Bi. = Binary.
\end{table}

\paragraph{Overall Precision and Recall on FinQA+TATQA.} The precision and recall metrics across models are given in Table~\ref{tab:finqa_precision} and Table~\ref{tab:finqa_recall}, respectively. We observe that both Phi4-mini-8k\textsuperscript{*} and Phi4-mini-36k\textsuperscript{*} perform poorly in numerical error detection. In contrast, their larger counterparts—Phi4-8k\textsuperscript{*} and Phi4-36k\textsuperscript{*}—demonstrate substantial improvements. For instance, precision increases from 29.3 to 92.1 when switching from Phi4-mini-8k\textsuperscript{*} to Phi4-8k\textsuperscript{*}. However, we also note that recall scores for the Phi-4 models tend to lag behind their precision scores, indicating that while they are highly accurate when predicting numerical errors, they may still miss a notable proportion of such cases.

\begin{table}[htbp]
\centering
\caption{Detection Performance (Precision) on FinQA+TATQA}
\begin{tabular}{lcccccccc}
\toprule
\textbf{Editor} & \textbf{Num.} & \textbf{Temp.} & \textbf{Ent.} & \textbf{Rel.} & \textbf{Con.} & \textbf{Unv.} & \textbf{Ov.} & \textbf{Bi.} \\
\midrule
GPT-4.1 mini & 14.4 & 21.7 & 17.8 & 57.1 & 53.8 & 92.3 & 37.9 & 64.3 \\
o3     &  37.2&  \underline{89.9}&  66.7&  39.5&  66.7&  \underline{93.8} &  68.5& 86.5\\
Phi4-mini-8k$^*$  & 29.3 & 51.2 & 37.8 & 40.7 & 70.0 & 74.5 & 63.3 &  84.1 \\
Phi4-mini-36k$^*$ & 46.6 & 76.2 & 37.5 & 38.2 & 78.4 & 82.1 & 71.3 &  87.6 \\
Phi4-8k$^*$       & \underline{92.1} & 88.5 & \underline{73.8} & \textbf{95.5} & \textbf{93.0} & \textbf{97.7} & \underline{93.0} & \textbf{98.3} \\
Phi4-36k$^*$ & \textbf{94.2} & \textbf{95.1} & \textbf{89.7} & \underline{89.7} & \textbf{93.0} & 93.4 & \textbf{95.0} &  \textbf{98.3} \\
QWEN3-4b-36k$^*$     & 50.0 & 62.7 & 47.8 & 42.9 & 79.2 & 82.1 & 72.1 & \underline{91.5} \\
QWEN3-14b-36k$^*$     & 62.7 & 67.0 & 56.1 & 41.7 & \underline{84.7} & 87.8 & 76.4 & 90.3 \\
\bottomrule
\end{tabular}
\label{tab:finqa_precision}
\\\footnotesize * are fine-tuned models.
\\\footnotesize Num. = Numerical, Tem. = Temporal, Ent. = Entity, Rel. = Relation, Con. = Contradictory, Unv. = Unverifiable, Ov. = Overall, Bi. = Binary.
\end{table}

\begin{table}[htbp]
\centering
\caption{Detection Performance (Recall) on FinQA+TATQA}
\begin{tabular}{lcccccccc}
\toprule
\textbf{Editor} & \textbf{Num.} & \textbf{Temp.} & \textbf{Ent.} & \textbf{Rel.} & \textbf{Con.} & \textbf{Unv.} & \textbf{Ov.} & \textbf{Bi.} \\
\midrule
GPT-4.1 mini & 63.3 & 28.4 & 22.2 & 53.3 & 13.5 & 27.3 & 58.4 & \textbf{98.4} \\
o3     &  \textbf{87.1}&  78.5&  61.1&  56.7&  18.9&  68.2&  75.6& 94.5\\
Phi4-mini-8k$^*$  & 28.8 & 52.5 & 47.2 & 36.7 & 71.0 & 77.8 & 64.4 & 83.6 \\
Phi4-mini-36k$^*$ & 56.7 & 58.5 & 42.9 & 43.3 & 90.6 & 71.1 & 72.7 & 89.1 \\
Phi4-8k$^*$       & 58.3 & \underline{83.1} & \underline{86.1} & \underline{70.0} & \textbf{97.1} & \textbf{95.6} & \underline{86.9} & 95.1 \\
Phi4-36k$^*$ & \underline{79.0} & \textbf{91.7} & \textbf{94.6} & \textbf{86.7} & \underline{95.7} & \textbf{95.6} & \textbf{92.7} & \underline{96.7} \\
QWEN3-4b-36k$^*$     & 46.6 & 55.3 & 62.9 & 62.1 & 86.4 & 71.1 & 72.7 & 88.0 \\
QWEN3-14b-36k$^*$     & 53.3 & 77.8 & 63.9 & 50.0 & 88.4 & \underline{80.0} & 79.0 & 91.8 \\
\bottomrule
\end{tabular}
\label{tab:finqa_recall}
\\\footnotesize * are fine-tuned models.
\\\footnotesize Num. = Numerical, Tem. = Temporal, Ent. = Entity, Rel. = Relation, Con. = Contradictory, Unv. = Unverifiable, Ov. = Overall, Bi. = Binary.
\end{table}

\subsection{Prompt for Baseline Models}\label{app:prompt}

Below is the prompt for detecting and editing hallucinations given the reference and passage as input arguments. This prompt is used for the evaluation of FinQA+TATQA datasets across all the baseline models.

\begin{tcolorbox}[title=Prompt for Baseline Models, colback=gray!5, colframe=black!80, sharp corners, breakable]
\begin{verbatim}
Given a passage with factual errors, identify any <numerical>, <temporal>, 
<entity>, <relation>, <contradictory>, or <unverifiable> errors in the passage 
and add edits for <numerical>, <temporal>, <entity> and <relation> errors by 
inserting additional <mark></mark> or <delete></delete> tags to mark and 
delete. If there are no errors, return the passage with no tags. Any changes 
to the original passage should be marked in <> tags. Below are the error 
definitions followed by examples of what you need to follow.

Definitions:

1. numerical errors (<numerical>): an incorrect calculation, estimation, or 
interpretation of numerical data such as percentages, growth rates, totals, 
differences, or ratios. These errors can arise from misapplying formulas, 
misreading data, rounding incorrectly, or failing to consider time periods or 
units.
2. temporal errors (<temporal>): incorrect reference or use of figures from 
the wrong time period. These errors typically arise from misinterpreting the 
reference's temporal context, such as year-over-year comparisons or quarter-
specific data.
3. entity errors (<entity>): a small part of a sentence, often an entity 
(e.g., location name), is incorrect (usually 1-3 words). Entity errors often
involve noun phrases or nouns.
4. relational error (<relation>): a sentence is partially incorrect as a small
part (usually 1 - 3 words). Relational errors often involve verbs and are 
often the opposite of what it should be.
5. contradictory sentence error (<contradictory>): a sentence where the entire
sentence is contradicted by the given reference, meaning the sentence can be
proven false due to a contradiction with information in the reference provided.
6. unverifiable sentence (<unverifiable>): a sentence where the whole sentence 
or phrase is unlikely to be factually grounded although it can be true, and 
the sentence cannot be confirmed nor denied using the reference given or 
internet search, it is often something personal or private and hence cannot 
be confirmed.

Follow the given example exactly, your task is to create the edited completion
with error tags <>:

Passage: Acme Corp's revenue reached $1.4 billion in Q4 2023, a 18% increase
compared to the same quarter in 2021.The company posted a net income of $450
million, up from $390 million the year before. Acme attributed the improved
financial performance to strong demand for its AI division and successful
restructuring efforts carried out earlier in the year.

Reference: In Q4 2023, Acme Corp reported a revenue of $2.4 billion, marking a
12% increase from Q4 2022. Net income for the quarter stood at $450 million, 
up from $390 million in the previous year. The company attributed this growth 
to increased demand in its cloud services division and operational 
efficiencies gained through restructuring efforts implemented in early 2023.

Edited: Acme Corp's revenue reached <numerical><delete>$1.4</delete><mark>
$2.4</mark></numerical> billion in Q4 2023, a <numerical><delete>18%</delete>
<mark>12%</mark></numerical> increase compared to the same quarter in 
<temporal><delete>2021</delete><mark>2022</mark></temporal>. The company 
posted a net income of $450 million, up from $390 million the year before. 
Acme attributed the improved financial performance to strong demand for its
<entity><delete>AI division</delete><mark>cloud services</mark></entity> and
successful restructuring efforts carried out earlier in the year.
<unverifiable>The restructuring was merely a PR move and had no financial 
impact.</unverifiable>

Now detect errors and include edits in the following passage like done in the 
example above. Include error tags <> for ANYTHING YOU CHANGE IN THE ORIGINAL 
PASSAGE.

Passage: [PASSAGE_TO_VERIFY]
Reference: [REFERENCE]

Return valid JSON in the following format:
{Edited: paragraph with inserted errors}
\end{verbatim}
\end{tcolorbox}

\section{Limitations}\label{app:limit}

Our experiments on hallucination detection and editing using SLMs demonstrate notable improvements over LLMs, while also revealing certain limitations. First, our current evaluation relies solely on language model-generated synthetic data; future work will extend this framework to real-world use cases to assess its practical applicability. Second, despite the observed performance gains, the underlying mechanisms that enable our framework to outperform baselines remain insufficiently understood. To address this, we plan to investigate the internal behavior of the system and provide mechanistic interpretability of the fine-tuned models in comparison to the baselines—an essential step for deployment in high-stakes, regulated domains.

\end{document}